\documentclass[letter, 10 pt, conference]{ieeeconf}  

\IEEEoverridecommandlockouts 
\usepackage[english]{babel}
\usepackage[utf8]{inputenc}
\usepackage{times} 
\usepackage{cite}

\usepackage{hyperref}
\usepackage[nolist]{acronym}

\usepackage{multirow}

\usepackage{siunitx}
\usepackage{graphicx} 
\usepackage{epsfig} 
\usepackage{pgfplots}
\usepackage{caption}
\usepackage[font=footnotesize]{subcaption}
\usepackage{pgfplots}

\usepackage{float}

\usepackage[export]{adjustbox}

\usepackage{mathtools}
\usepackage{nicefrac}
\usepackage{amsmath}
\usepackage{amssymb}
\usepackage{bm} 

\DeclareMathOperator*{\minimize}{minimize}

\usepackage{etoolbox}
\makeatletter%
\AfterPreamble{%
	\usepackage{hyperref}%
	\let\oldhypertarget\hypertarget%
	\renewcommand{\hypertarget}[2]{%
		\oldhypertarget{#1}{#2}%
		\protected@write\@mainaux{}{%
			\string\expandafter\string\gdef%
			\string\csname\string\detokenize{#1}\string\endcsname{#2}%
		}%
	}%
	\newcommand{\myhyperlink}[1]{%
		\hyperlink{#1}{\csname #1\endcsname}%
	}%
}
\makeatother%

\newcounter{Remark}
\newcommand{\displayRemarks}[2][]{%
	\stepcounter{Remark}%
	\textit{Remark~}\hypertarget{#1}{\theRemark}\textit{~(#2)}%
}

\newcounter{Problem}
\usepackage{algorithm}
\usepackage[noend]{algpseudocode}

\makeatletter
\def\BState{\State\hskip-\ALG@thistlm}
\makeatother

\usepackage{tikz}
\pgfdeclarelayer{foreground}
\pgfsetlayers{background, main, foreground}
\usetikzlibrary{quotes, angles, backgrounds, arrows, automata, shapes, positioning, calc, through, spy, decorations.pathreplacing, decorations.markings, arrows.meta, automata, petri}

\tikzset{
    imglabel/.style={
      rectangle,
      inner sep=2pt,
      text=black,
      minimum height=1em,
      text centered,
      fill=white,
      fill opacity=1.0,
      text opacity=1,
      anchor=south west,
    },
  }
\tikzset{
	state/.style={
		rectangle,
		draw=black, very thick,
		minimum height=1.0em,
		text centered,
	},
}
\tikzset{
  on each segment/.style={
    decorate,
    decoration={
      show path construction,
      moveto code={},
      lineto code={
        \path [#1]
        (\tikzinputsegmentfirst) -- (\tikzinputsegmentlast);
      },
      curveto code={
        \path [#1] (\tikzinputsegmentfirst)
        .. controls
        (\tikzinputsegmentsupporta) and (\tikzinputsegmentsupportb)
        ..
        (\tikzinputsegmentlast);
      },
      closepath code={
        \path [#1]
        (\tikzinputsegmentfirst) -- (\tikzinputsegmentlast);
      },
    },
  },
  mid arrow/.style={postaction={decorate,decoration={
        markings,
        mark=at position .5 with {\arrow[#1]{stealth}}
      }}},
}

\graphicspath{{./figures/}}

\newcommand\copyrighttext{%
	\small \begin{center} \color{red} \textcopyright\,2022 IEEE. Personal use of this material is permitted. Permission from IEEE must be obtained for all other uses, in any current or future media, including reprinting/republishing this material for advertising or promotional purposes, creating new collective works, for resale or redistribution to servers or lists, or reuse of any copyrighted component of this work in other works. \end{center}}
\newcommand\copyrightnotice{%
	\begin{tikzpicture}[remember picture,overlay]
	\node[anchor=south,yshift=25.6cm] at (current page.south) 
	{\color{red}\fbox{\parbox{\dimexpr\textwidth-\fboxsep-\fboxrule\relax}{\copyrighttext}}};
	\end{tikzpicture}%
}

\title{\copyrightnotice \LARGE \bf A Perception-Aware NMPC for Vision-Based Target Tracking and Collision Avoidance with a Multi-Rotor UAV} 

\author{Andriy Dmytruk$^{1}$, Giuseppe Silano$^{1}$, Davide Bicego$^{2}$, Daniel Bonilla Licea$^{1}$, and Martin Saska$^{1}$  
    \thanks{This work was partially funded by the European Union's Horizon 2020 research and innovation programme AERIAL-CORE under grant agreement no. 871479, by Czech Science Foundation (GAČR) under research project no. 20-10280S, by CTU grant no. SGS20/174/OHK3/3T/13, and by OP VVV funded project CZ.02.1.01/0.0/0.0/16 019/0000765 ``Research Center for Informatics".}
    \thanks{$^{1}$Andriy Dmytruk, Giuseppe Silano, Daniel Bonilla Licea, and Martin Saska are with the Faculty of Electrical Engineering, Czech Technical University in Prague, Czech Republic, (email: {\tt\small \{dmytrand, silangiu, bonildan, saskam1\}@fel.cvut.cz).}} 
    \thanks{$^{2}$Davide Bicego is with the Robotics and Mechatronics group, Faculty of Electrical Engineering, Mathematics \& Computer Science, University of Twente, Netherlands (email: {\tt\small d.bicego@utwente.nl).}}
}

\begin{document}

\maketitle
\thispagestyle{empty} 
\pagestyle{empty} 


\begin{acronym}
    \acro{CCW}[CCW]{Counter-ClockWise}
    \acro{CW}[CW]{ClockWise}
    \acro{CoM}[CoM]{Center of Mass}
    \acro{FoV}[FoV]{Field of View}
    \acro{GTMR}[GTMR]{Generically Tilted Multi-Rotor}
    \acro{MPC}[MPC]{Model Predictive Control}
    \acro{NLP}[NLP]{Nonlinear Programming}
    \acro{NMPC}[NMPC]{Nonlinear Model Predictive Control}
    \acro{QP}[QP]{Quadratic Programming}
    \acro{ROS}[ROS]{Robot Operating System}
	\acro{UAV}[UAV]{Unmanned Aerial Vehicle}
	\acro{wrt}[w.r.t.]{with respect to}
\end{acronym}



\begin{abstract}

A perception-aware~\acf{NMPC} strategy aimed at performing vision-based target tracking and collision avoidance with a multi-rotor aerial vehicle is presented in this paper. The proposed control strategy considers both realistic actuation limits at the torque level and visual perception constraints to enforce the visibility coverage of a target while complying with the mission objectives. Furthermore, the approach allows to safely navigate in a workspace area populated by dynamic obstacles with a ballistic motion. The formulation is meant to be generic and set upon a large class of multi-rotor vehicles that covers both coplanar designs like quadrotors as well as fully-actuated platforms with tilted propellers. The feasibility and effectiveness of the control strategy are demonstrated via closed-loop simulations achieved in MATLAB.

\end{abstract}






\section{Introduction}
\label{sec:introduction}

In the last decade, the research and commercial interest on~\acp{UAV} has exploded as demonstrated by the growing number of applications, such as infrastructure monitoring~\cite{Silano2021RAL}, aerial filming~\cite{Kratky2021RAL}, surveillance and search and rescue missions~\cite{Petracek2021RAL} and wireless communications~\cite{Licea2021EUSIPCO}. Most of these applications require~\acp{UAV} with visual sensors and with certain autonomy that enable them to rapidly react in dynamic environments to successfully accomplish their mission. To achieve this objective, perception constraints must be integrated within the~\acp{UAV} control framework.

Numerous control strategies have been proposed to perform this integration. Among these, the~\ac{MPC} scheme has been proven to be a promising solution to control the~\ac{UAV} motion while complying with its dynamics and multiple heterogeneous constraints, such as those coming from the visual part~\cite{JacquetRAL2022, JacquetRAL2021, Falanga2018IROS}. Specifically,~\acf{NMPC} has resulted particularly suitable to control aerial vehicles when their agility is essential for the specific application and must be exploited at the best~\cite{Kamel2017IFAC}. In this formulation, nonlinear models and constraints are used to predict the~\ac{UAV} dynamics alongside with the mission objectives. Recent works explored extensions of this formulation using adaptive~\cite{Hanover2021adaptive} and data-driven~\cite{Torrente2021RAL} approaches.

Relevant advances in the computational capabilities of modern computers and improvements in the algorithms efficiency~\cite{Andersoon2019, Ferreau2014, Verschueren2021} have made it possible to manage the high-computational load and stringent real-time requirements to solve these problems. Several software frameworks~\cite{Sathya2018ECC, Pantelis2020IFAC, Chen2019ECC} have been released over the years to facilitate modeling, control design and simulation for a broad class of~\ac{MPC} applications. 

In this context, various works have investigated~\ac{NMPC} strategies considering perception-objectives~\cite{Paneque2022RAL, Kamel2017IFAC, Bicego2020JINT}. A common problem studied in those works is the tracking of a target subject to certain visibility constraints. In most cases,~\ac{NMPC} is used in the outer loop of a cascaded architecture to provide a reference trajectory to an inner loop pose tracking controller~\cite{Paneque2022RAL}. This approach allows to attain the perception-objectives, but it can cause problems as the~\ac{NMPC} generator does not consider the limitations posed by the low-level controller~\cite{Kamel2017IFAC, Bicego2020JINT}. As a consequence, the generated trajectory could violate the actuation limitations of the~\ac{UAV} resulting in an unfeasible solution.

To overcome this limitation,~\ac{NMPC} can be used to combine trajectory generation, subject to visibility and collision avoidance constraints, and trajectory tracking, subject to actuation constraints, in a single optimization problem, as shown in~\cite{JacquetRAL2022, Penin2018RAL, Falanga2018IROS}. This allows to account for partially target occlusions that may appear as the camera moves, and physical constraints, such as feasibility and collision avoidance constraints, that could jeopardize the mission accomplishment. These approaches leverage on the intrinsic capability of the optimization framework to include inequality and equality constraints of various semantics. Therefore, the so-formulated problem allows to keep tracking of the target feature, preventing critical configurations related to~\ac{FoV} constraints, while taking into account the limits imposed by actuators.

Such single control layer architectures with perception and actuation constraints have been investigated also in~\cite{JacquetRAL2021, JacquetRAL2022}. In these works, the authors make use of generic dynamic models which can represent a large class of multi-rotor~\acp{UAV}~\cite{Bicego2020JINT}. This allows to formulate the problem for both standard coplanar under-actuated vehicles as well as for fully-actuated platforms with tilted propellers characterized by a fixed geometry (i.e., propellers' orientation is fixed). However, collision avoidance tasks~\ac{wrt} multiple dynamic obstacles are not explicitly taken into account.


Following this line of research, a~\ac{NMPC} architecture for vision-driven target tracking and collision avoidance that considers both visual perception constraints and physical actuation limitations of a broad class of multi-rotor platforms is proposed. In particular, the problem where a~\ac{GTMR}~\cite{Ryll2019IJRR, Michieletto2018TRO} equipped with a visual sensor is required to track the trajectory of a moving target while keeping it in the camera~\ac{FoV} is considered. Meanwhile, the~\ac{GTMR} is required to safely navigate in a workspace area populated by dynamic obstacles characterized by a ballistic motion. The initial position of the obstacles and their motion are assumed to be known in the whole prediction horizon of the~\ac{NMPC}. Such a scenario aims at simulating the motion of balls thrown by a person\footnote{\url{https://youtu.be/w2itwFJCgFQ?t=531}}. The visual perception objectives are implemented as hard constraints, while the obstacle avoidance is enforced by soft constraints. This approach compensates for the lack of knowledge on the target motion by exploiting those of the obstacles to relax the optimization problem. Soft constraints ensure continuity of the solution when constraints may arise unfeasibility issues. 

The paper is organized as follows. Section~\ref{sec:systemModeling} describes the model of a~\ac{GTMR} and a generic visual sensor considered equipping the vehicle. Section~\ref{sec:optimalControlProblemFormulation} presents the optimal control problem formulation, including the equality and inequality constraints for the collision avoidance and visual-target tracking. Section~\ref{sec:simulationResults} reports the simulations results achieved in MATLAB, which are used to demonstrate the validity of the proposed approach. Finally, Section~\ref{sec:conclusions} concludes the paper.

\section{System Modeling}
\label{sec:systemModeling}




\subsection{System dynamics}
\label{sec:systemDynamics}

\begin{figure}[tb]
    \centering
    \scalebox{1}{
    \begin{tikzpicture}
		\draw[-latex] (-3,-1.5) node[below]{$O_W$}  -- (-2,-1.5) node[below]{$\hat{x}_W$}; 
		\draw[-latex] (-3,-1.5) -- (-3,-0.5) node[left]{$\hat{z}_W$};; 
		\draw[-latex] (-3,-1.5) -- (-2.5,-1) node[above]{$\hat{y}_W$};; 
		
		\node (quadrotor) at (0, 0.75) {\includegraphics[scale=1.75]{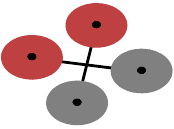}};
        \node (camera) at (2, -1.5) {\includegraphics[rotate=-10, scale=1.75]{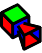}};
        
		\draw[-latex] (-0.175,0.35) arc (55:-15:0.35cm); 
		\draw[-latex] (0.975,1.00) arc (55:-15:0.35cm); 
		\draw[latex-] (0.40,1.75) arc (-25:-120:0.30cm); 
		\draw[latex-] (-0.75,1.15) arc (-25:-120:0.30cm); 
		
		\draw[-latex] (-0.175,0.05) coordinate -- (-0.175,0.55); 
		\draw (-0.075,0.50) node[below right]{$\Omega_1$};
		\draw[-latex] (0.975,0.70) coordinate -- (0.975,1.20);
		\draw (1.15,1.10) node[below right]{$\Omega_2$};
		\draw[-latex] (0.175,1.45) coordinate -- (0.175,1.95);
		\draw (0.175,1.95) node[right]{$\Omega_3$};
		\draw[-latex] (-0.975,0.95) coordinate -- (-0.975,1.45);
		\draw (-0.975,1.25) node[left]{$\Omega_4$};
		
		\draw[-latex] (0,0.75)  -- (1.25,1.25) node[above]{$\hat{y}_B$}; 
		\draw[-latex] (0,0.75) -- (0,2.00) node[left]{$\hat{z}_B$};; 
		\draw[-latex] (0,0.75) -- (1,0) node[below right]{$\hat{x}_B$};; 
		\draw (-0.35,0.85) node[below]{$O_B$}; 
		
		\draw[-latex] (2, -1.5) -- (2.7, -3.3); 
		\draw[-latex] (2.25, -1.5) node[right]{$\hat{z}_C$};
		\draw[-latex] (2, -1.5) -- (2, -2.5) node[left]{$\hat{y}_C$};; 
		\draw[-latex] (2, -1.5)  -- (1, -1.8) node[below]{$\hat{x}_C$}; 
		\draw (1.65, -1.6) node[below]{$O_C$}; 
		
		\draw[dashed, thin, gray] (2, -1.5) -- (3.2, -3.3); 
		\draw[dashed, thin, gray] (2, -1.5) -- (2.2, -3.3); 
		\draw[dashed, stealth-stealth, thin, green] (2.2, -3.2) --  (3, -3) node[below, text centered]{\scriptsize{$\alpha_h$}}; 
		\draw[dashed, thin, gray] (2, -1.5) -- (2.7, -2.90); 
		\draw[dashed, thin, gray] (2, -1.5) -- (2.7, -3.75); 
		\draw[dashed, stealth-stealth, thin, red] (2.4, -2.8) -- node[right, text centered]{\scriptsize{$\alpha_v$}} (2.4, -2.50); 
		\draw[draw=black] (3.2, -2.75) -- (3.2, -3.65)  -- (2.2, -3.9) -- (2.2, -3.0) -- (3.2, -2.75);
		\node (point1) at (2.7, -3.3) [circle,fill,draw,inner sep=0pt,minimum size=3pt, text centered]{};
		\draw (2.7, -3.5) node[text centered, rotate=10]{$\prescript{C}{}{\mathbf{p}_M}$};
		\node (point2) at (3.8, -3.8) [circle,fill,draw,inner sep=0pt,minimum size=3pt, text centered]{};
		\draw (3.8, -4.0) node[text centered, rotate=0]{$\mathbf{p}_M$};
		\draw[dashed, gray] (point1) -- (point2);
		\draw[-latex, dashed, gray] (1.2, -3.5) -- (2.1, -3.5); 
		\draw (1.2, -3.9) node[above, text width=5em]{Image\\Plane};
		
		\draw[dashed, thin] (-3,-1.5) -- (0,0.75); 
		\draw (-1.15,-0.9) node[above, rotate=40]{$\mathbf{R}(\mathbf{q})$}; 
		\draw[dashed, thin] (0, 0.75) -- (2, -1.5); 
		\draw (0.75, -0.95) node[above, rotate=-47.5]{$\mathbf{R}(\mathbf{q_C})$}; 
		
    \end{tikzpicture}
    }
    \caption{A schematic representation of a~\ac{GTMR} model equipped with a generic visual sensor in the case of four propellers ($n=4$) and coplanar orientation.}
    \label{fig:sampleGenericallyTiltedMultiRotor}
\end{figure}
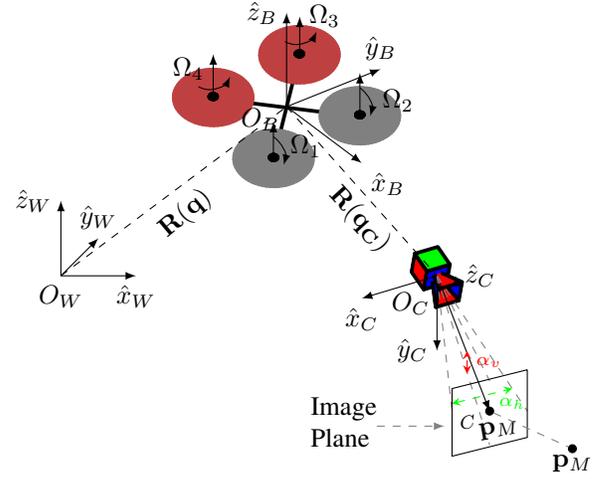

Let us consider a~\ac{GTMR} model~\cite{Michieletto2018TRO}, composed of a rigid body and $n \in \mathbb{R}_{>0}$ propellers that spin about a generically oriented axis with fixed but arbitrary orientations. The relative propeller orientation, jointly with the number $n$ of rotors, determine whether the~\ac{GTMR} is an under-actuated or a fully-actuated platform~\cite{Ryll2019IJRR}. A schematic representation of the system is reported in Fig.~\ref{fig:sampleGenericallyTiltedMultiRotor}.

Let us denote with $\mathcal{F}_W$ and $\mathcal{F}_B$ the \textit{world frame} and \textit{body frame} reference systems, respectively. The body frame is attached to the~\ac{GTMR} so that the origin of the frame $O_B$ coincides with the~\ac{CoM} of the vehicle. The position of the origin $O_B$ of the body frame $\mathcal{F}_B$~\ac{wrt} the world frame $\mathcal{F}_W$ is denoted with $\mathbf{p} \in \mathbb{R}^3$, while the unit quaternion representing the rotation from the body frame $\mathcal{F}_B$ to the world frame $\mathcal{F}_W$ is denoted as  $\mathbf{q} \in \mathbb{S}^3$. The angular velocity of the~\ac{GTMR} in $\mathcal{F}_B$~\ac{wrt} $\mathcal{F}_W$, expressed in $\mathcal{F}_B$, is denoted with $\bm{\omega} \in \mathbb{R}^3$, whereas the linear velocity of $O_B$ in $\mathcal{F}_W$ is denoted by $\mathbf{v} = \dot{\mathbf{p}} \in \mathbb{R}^3$.

The $i$-th propeller spinning velocity $\Omega_i \in \mathbb{R}_{\geq 0}$, with $i = \{1, 2, \dots, n \}$, represents the controllable input variable of the system, i.e., $\mathbf{u} = \bm{\Omega}$ and $\bm{\Omega} = [ \Omega_1 \cdots \Omega_n ]^\top \in \mathbb{R}^n_{\geq 0}$. While rotating, each propeller exerts a thrust force $\mathbf{f}_i \in \mathbb{R}^3$ oriented along the axis perpendicular to the plane spanned by the propeller. Following the right-hand convention, this force generates a drag momentum $\bm{\tau}_i \in \mathbb{R}^3$ oriented as the angular velocity vector $\bm{\omega}$ in case of clockwise rotation, and opposite in case of counter-clockwise rotation. The sum of all forces $\mathbf{f}_i$ coincides with the control force $\mathbf{f}_c  \in \mathbb{R}^3$ exerted at the platform~\ac{CoM}, while the control momentum $\bm{\tau}_c \in \mathbb{R}^3$ is the sum of the momentum contributions $\bm{\tau}_i$ due to both the thrust forces and the drag momenta. Further details on the model derivation are available in~\cite{Michieletto2018TRO}.

After neglecting second order effects and using the Newton-Euler approach, the~\ac{GTMR} dynamics can be approximated by the set of equations
\begin{equation}\label{eq:modelEquations}
\left\{
	\begin{array}{l}
    \dot{\mathbf{p}} = \mathbf{v} \\
    \dot{\mathbf{q}} = \frac{1}{2} \mathbf{q} \circ \begin{bmatrix} 0 \\ \bm{\omega} \end{bmatrix} \\
    m \dot{\mathbf{v}} = -mg \mathbf{e}_3 + \mathbf{R}(\mathbf{q}) \mathbf{F} \mathbf{u} \\
    \mathbf{J} \dot{\bm{\omega}} = - \bm{\omega} \times \mathbf{J} \bm{\omega} + \mathbf{M} \mathbf{u}
	\end{array}
\right.,
\end{equation}
where $m \in \mathbb{R}_{>0}$ and $g \in \mathbb{R}_{>0}$ denote the~\ac{GTMR} mass and the gravitational constant, respectively, $\mathbf{e}_m \in \mathbb{R}^3$, with $m \in \{ 1, 2, 3 \}$, represents the $m$-th column of the identity matrix $\mathbf{I}_3 \in \mathbb{R}^{3 \times 3}$, and $\mathbf{J} \in \mathbb{R}^{3 \times 3}$ is the positive definite constant~\ac{GTMR} inertia matrix in $\mathcal{F}_B$. The symbols $\circ$ and $\times$ denote the quaternion product and the vector cross product operations, respectively. $\mathbf{R}$ is the rotation matrix from the body $\mathcal{F}_B$ to the world frame $\mathcal{F}_W$. Finally, $\mathbf{F} \in \mathbb{R}^{3 \times n}$ and $\mathbf{M} \in \mathbb{R}^{3 \times n}$ are the forces and momenta allocation matrices, respectively, mapping the vector of forces produced by each $i$-th propeller to the total force and momentum acting on the vehicle~\ac{CoM}~\cite{Michieletto2018TRO}. The model~\eqref{eq:modelEquations} describes a nonlinear dynamic system $\dot{\mathbf{x}} = \mathbf{f} ( \mathbf{x}, \mathbf{u} )$, with state $\mathbf{x} = [ \mathbf{p}^\top \mathbf{q}^\top \mathbf{v}^\top \bm{\omega}^\top  ]^\top \in \mathbb{R}^3 \times \mathbb{S}^3 \times \mathbb{R}^6$ and control input $\mathbf{u} = \bm{\Omega}$. 

To account for the limited bandwidth of the control action variable $\mathbf{u}$ (i.e., maximum derivative of the propeller spinning velocity value), smoothness has to be enforced. In this regard, it is convenient to extend model~\eqref{eq:modelEquations} assuming as new control input variable $\bar{\mathbf{u}}=\dot{\mathbf{u}}$, i.e., the time derivative of the propeller spinning velocity $\dot{\bm{\Omega}} \in \mathbb{R}^n$, including $\mathbf{u}$ among the system state variables $\mathbf{x}$~\cite{Bicego2020JINT}. Therefore, the system model~\eqref{eq:modelEquations} can be rewritten as $\dot{\bar{\mathbf{x}}} = \mathbf{f}(\bar{\mathbf{x}}, \bar{\mathbf{u}})$, where $\bar{\mathbf{x}} = [ \mathbf{p}^\top \mathbf{q}^\top \mathbf{v}^\top \bm{\omega}^\top \mathbf{u}^\top ]^\top$ and $\bar{\mathbf{u}} = \dot{\mathbf{u}}$. In this way, realistic physical limitations on the actuators, i.e., lower and upper bounds on the rotor accelerations, which lie at the same differential level of the motor torques, can be considered within the system modeling~\eqref{eq:modelEquations} and in the control problem formulation (Section~\ref{sec:optimalControlProblemFormulation}). Thus, following the identification process in~\cite{Bicego2020JINT}, the actuator bounds can be expressed as
\begin{subequations}\label{eq:constraintsVehicle}
    \begin{align}
        \underline{\bm{\gamma}} \leq \mathbf{u} \leq \bar{\bm{\gamma}}, \\
        \underline{\dot{\bm{\gamma}}} \leq \bar{\mathbf{u}} \leq \bar{\dot{\bm{\gamma}}},
    \end{align}
\end{subequations}
where $\underline{\bm{\gamma}}$ and $\bar{\bm{\gamma}}$ represent the minimum and maximum propeller spinning velocity values, respectively, while $\underline{\dot{\bm{\gamma}}}$ and $\bar{\dot{\bm{\gamma}}}$ are the minimum and maximum propeller spinning acceleration values, respectively, which can depend on the rotor velocity.




\subsection{Generic sensor model}
\label{sec:genericSensorModel}


For the sake of generality of the proposed control strategy, the~\ac{GTMR} is considered to be equipped with a generic visual sensor able to retrieve the position of the target on the camera image plane. Such an assumption allows synthesizing an optimal control problem independent from the type of sensor used to detect the target. The visual sensor reference frame is denoted as $\mathcal{F}_C$ with optical axis $\hat{z}_C$, as depicted in Fig.~\ref{fig:sampleGenericallyTiltedMultiRotor}. Also, the visual sensor is assumed to be rigidly attached to the~\ac{GTMR} body and its pose ($\mathbf{p}_C$, $\mathbf{q}_C$) to be known. In other words, the position ($\mathbf{p}_C$) and orientation ($\mathbf{q}_C$) of the origin $O_C$ of the camera frame $\mathcal{F}_C$~\ac{wrt} the body frame $\mathcal{F}_B$ are assumed to be fixed and known. Hence, the transformation between the visual sensor reference frame $\mathcal{F}_C$ and the~\ac{GTMR} body frame $\mathcal{F}_B$ is also known. Besides, a \textit{pinhole camera model} is used to describe the mathematical relationship between the position of the target $\mathbf{p}_M$ in $\mathcal{F}_W$ and its projection $\prescript{C}{}{\mathbf{p}_M}$ onto the image plane $\mathcal{F}_C$ while neglecting possible distortions on the image produced by lenses~\cite{Falanga2018IROS}. Therefore, given a generic point $\mathbf{p}_M = [ x_M \, y_M \, z_M ]^\top$ in the world frame $\mathcal{F}_W$, its projection onto the camera frame $\mathcal{F}_C$ is denoted as $\prescript{C}{}{\mathbf{p}_M}= [ \prescript{C}{}{x_M} \, \prescript{C}{}{y_M} \, \prescript{C}{}{z_M} ]^\top$. Under the assumptions of a pyramidal~\ac{FoV} (see Fig.~\ref{fig:sampleGenericallyTiltedMultiRotor}), the constraints to enforce the visibility coverage of the target position $\mathbf{p}_M$ can be defined as
\begin{subequations}
    \begin{align}
        \lvert \prescript{C}{}{x_M} / \prescript{C}{}{y_M} \rvert &\leq \tan{\alpha}_h, \\
        \lvert \prescript{C}{}{y_M} / \prescript{C}{}{z_M} \rvert &\leq \tan{\alpha}_v,  
    \end{align}
\end{subequations}
where $\alpha_h$ and $\alpha_v$ denote the horizontal and the vertical angles of the pyramidal~\ac{FoV}, respectively, as depicted in Fig.~\ref{fig:sampleGenericallyTiltedMultiRotor}.





\section{Optimal Control Problem Formulation}
\label{sec:optimalControlProblemFormulation}



A~\ac{GTMR} equipped with a generic visual sensor has to track the motion of a target, whose motion is unknown, and maintain it in the camera image plane. Meanwhile, the multi-rotor is required to avoid dynamic obstacles populating the workspace. The proposed control setup has to comply with the platform physical limitations and vision constraints, while fulfilling the assigned mission at the best. The proposed framework is stated for one sensor and a single target feature. However, such an assumption does not preclude to extend the framework to consider more targets as in~\cite{JacquetRAL2022}. The motion of the target is such as to ensure that the~\ac{GTMR} can track it.

The following sections deal with describing the collision constraints applied to the system, the objective function, and the~\ac{NMPC} optimal formulation.



\subsection{Collision avoidance}
\label{sec:collissionAovidance}


As stated in Sec.~\ref{sec:optimalControlProblemFormulation}, the~\ac{GTMR} is required to maintain the target within the camera~\ac{FoV} while avoiding obstacles in the workspace area. The obstacles' motion is assumed to be known in the whole prediction horizon of the~\ac{NMPC}. Besides, a point-mass model is used to approximate the~\ac{GTMR} body and the obstacle sizes leaving out the actual sizes of the robot and those of the obstacles from the collision avoidance requirements.  

Let us denote with $\mathbf{p}_{\mathrm{o}_j} = [ x_{\mathrm{o}_j} \, y_{\mathrm{o}_j} \, z_{\mathrm{o}_j} ]^\top$ the position of the $j$-th obstacle in the world frame $\mathcal{F}_W$, with $j \in \{ 1, 2, \dots, O\}$ and $O \in \mathbb{N}_{>0}$ denoting the number of obstacles populating the workspace area. Therefore, the collision avoidance constraint is considered to be the square Euclidean distance defined as
\begin{equation}\label{eq:constraintDistance}
   \lVert \mathbf{p} - \mathbf{p}_{\mathrm{o}_j} \rVert^2 \geq \Gamma_j^2, 
\end{equation}
where $\Gamma_j \in \mathbb{R}_{>0}$ is the minimum distance value the multi-rotor has to maintain to avoid crashes with the obstacles.

%
%

\displayRemarks[RemarkSolvers]{Optimal Solvers}: Note that for optimization solvers whose algorithm is based on a Newton-type method, such as ACADOS~\cite{Verschueren2021}, PANOC~\cite{Sathya2018ECC}, and CasADi~\cite{Andersoon2019}, only one system linearization and~\ac{QP} problem are performed per time. Each~\ac{QP} problem corresponds to a linear approximation of the original~\ac{NLP} problem along a time-varying trajectory. In this configuration, squaring the Euclidean norm, as in~\eqref{eq:constraintDistance}, provides local trajectories that are over-conservative. In a case where the linearization points are far away from the boundaries of the linearized feasible set of the original~\ac{NLP}, the use of~\eqref{eq:constraintDistance} leads to linear constraints that are ``least" conservative. Further details can be found in~\cite{BarrosIFAC2020}.

\subsection{Objective function}
\label{sec:objectiveFunction}

A motion tracking problem is described as a minimization distance task between the vehicle state $\mathbf{x}$ and a reference motion $\mathbf{x}_\mathrm{d}$. Usually, a reference trajectory is expressed in both position and attitude, denoted as $(\mathbf{p}_\mathrm{d}, \mathbf{q}_\mathrm{d})$, and the corresponding first $(\mathbf{v}_\mathrm{d}, \bm{\omega}_\mathrm{d})$ and second $(\dot{\mathbf{v}}_\mathrm{d}, \dot{\bm{\omega}}_\mathrm{d})$ order time derivatives. These signals are sampled over the prediction horizon of the optimization problem and retrieved as output of a trajectory planner. Examples of these schemes can be found in~\cite{Falanga2018IROS, JacquetRAL2022, Penin2018RAL}.

The trajectory tracking error is defined as a weighted square Euclidean norm denoted as $\lVert \cdot \rVert^2_\mathrm{\mathbf{Q}}$, where $\mathbf{Q} \in \mathbb{R}_{\geq 0}$ is a diagonal weight matrix used as a tunable controller gain. Note that the Euclidean norm cannot be used to model the differences between the reference attitude $\mathbf{q}_\mathrm{d}$ and the drone attitude $\mathbf{q}$ due to the disambiguities introduced by the unit quaternion representation. Indeed, the $\mathbf{q}$ and $-\mathbf{q}$ unit quaternions represent the same attitude. 


To overcome this issue, the geodesic distance described in~\cite{JacquetRAL2021} was considered to model the difference between quaterions. Hence, given the unit quaternions $\mathbf{q}_1$ and $\mathbf{q}_2$, their distance can be calculated as $\lVert \log(\mathbf{q}_1 \circ \mathbf{q}^\star_2) \rVert$, where $\mathbf{q}^\star_\bullet$ represents the conjugate of the unit quaternion $\mathbf{q}_\bullet$. Note that, in order to improve the paper readability, with an abuse of notation the weighted Euclidean norm $\lVert \mathbf{q} - \mathbf{q}_\mathrm{d} \rVert^2_\mathrm{\mathbf{Q}}$ is used to denote the weighted attitude error between the unit quaternions describing the drone ($\mathbf{q}$) and reference attitudes ($\mathbf{q}_\mathrm{d}$), instead of referring to the weighted geodesic distance $\lVert \log(\mathbf{q}_1 \circ \mathbf{q}^\star_2) \rVert^2_\mathrm{\mathbf{Q}}$. Such a representation allows to compact the notation of the optimal control problem (Section~\ref{sec:optimalControlProblem}).

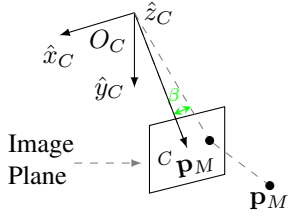
\begin{figure}[tb]
    \centering
    \begin{tikzpicture}
    
        \draw[dashed, thin, gray] (2, -1.5) -- (2.7, -3.3); 
		\draw[dashed, stealth-stealth, thin, green] (2.75, -2.75) --  (2.52, -2.85) node[above, text centered]{\scriptsize{$\beta$}}; 
    
		\draw[-latex] (2, -1.5)  -- (1, -1.8) node[below]{$\hat{x}_C$}; 
		\draw[-latex] (2, -1.5) node[right]{$\hat{z}_C$} -- (2.7, -3.3); 
		\draw[-latex] (2, -1.5) -- (2, -2.5) node[left]{$\hat{y}_C$};; 
		\draw (1.65, -1.6) node[below]{$O_C$}; 
		
		\draw[draw=black] (3.2, -2.75) -- (3.2, -3.65)  -- (2.2, -3.9) -- (2.2, -3.0) -- (3.2, -2.75);
		\node (point1) at (3.0, -3.2) [circle,fill,draw,inner sep=0pt,minimum size=3pt, text centered]{};
		\draw[dashed, gray] (2, -1.5) -- (point1);
		\draw (2.7, -3.5) node[text centered, rotate=10]{$\prescript{C}{}{\mathbf{p}_M}$};
		\node (point2) at (3.8, -3.8) [circle,fill,draw,inner sep=0pt,minimum size=3pt, text centered]{};
		\draw (3.8, -4.0) node[text centered, rotate=0]{$\mathbf{p}_M$};
		\draw[dashed, gray] (point1) -- (point2);
		\draw[-latex, dashed, gray] (1.2, -3.5) -- (2.1, -3.5); 
		\draw (1.2, -3.9) node[above, text width=5em]{Image\\Plane};
        
    \end{tikzpicture}
    \caption{A schematic representation of the angular position $\beta$ of an image point $\prescript{C}{}{\mathbf{p}_M}$.}
    \label{fig:representationBeta}
\end{figure}

Along with tracking errors, perception errors are also embedded into the objective function of the optimal control problem to maintain the target in the camera~\ac{FoV}. Following~\cite{Penin2018RAL}, the angular distance $\beta$ between the target position on the image plane ($\prescript{C}{}{\mathbf{p}_M}$) and the $z$-axis of the camera frame $\mathcal{F}_C$ is minimized. In addition, to avoid motion blur and smoothly guide the target tracking, the time derivative $\dot{\beta}$ is also accounted in the optimal problem formulation. Figure~\ref{fig:representationBeta} shows a schematic representation of the perception objectives. 

To make the problem resolution more efficient, the difference in angle and its derivative are replaced with the cosine function, i.e., $c\beta$ and $\dot{c\beta}$, where $c\bullet$ refers to the short notation for the cosine function. Besides, a weighted distance term $\lVert \text{dist}( \mathbf{p}, \mathbf{p}_M ) - \Upsilon \rVert^2_\mathbf{Q}$ enforces a constant distance $\Upsilon \in \mathbb{R}_{>0}$ between the drone ($\mathbf{p}$) and the target ($\mathbf{p}_M$) positions. The $\text{dist}(\bullet, \bullet)$ function refers to the Euclidean distance function between two points. 


Therefore, the objective function can be written as the weighted difference between the desired $\mathbf{y}_\mathrm{d}$ and the current $\mathbf{y} = \mathbf{h}(\bar{\mathbf{x}}, \bar{\mathbf{u}}, \mathbf{p}_M)$ output signals, i.e., $\lVert \mathbf{y} - \mathbf{y}_\mathrm{d} \rVert_\mathbf{\mathbf{Q}}$, as
\begin{subequations}
    \begin{align}
        \mathbf{y}_\mathrm{d} &= [\mathbf{p}_\mathrm{d}^\top \, \mathbf{q}_\mathrm{d}^\top \, \mathbf{v}_\mathrm{d}^\top \, \bm{\omega}_\mathrm{d}^\top \, \dot{\mathbf{v}}_\mathrm{d}^\top \, \dot{\bm{\omega}}_\mathrm{d}^\top 1 \,\, 0 \,\, \Upsilon ]^\top , \\
        %
        \mathbf{y} &= [\mathbf{p}^\top \, \mathbf{q}^\top \, \mathbf{v}^\top \, \bm{\omega}^\top \, \dot{\mathbf{v}}^\top \, \dot{\bm{\omega}}^\top \text{c}{\beta} \;\; \dot{\text{c}{\beta}} \;\; \text{dist}( \mathbf{p}, \mathbf{p}_M)]^\top .
    \end{align}
\end{subequations}



\subsection{Optimal control problem}
\label{sec:optimalControlProblem}

Thus, the optimal control problem over a prediction horizon of $N$ steps, where $N \in \mathbb{N}_{>0}$, considering the revised system model equations $\dot{\bar{\mathbf{x}}} = \mathbf{f}( \bar{\mathbf{x}}, \bar{\mathbf{u}} )$, can be formulated as a minimization problem per each time step $\mathbf{t}_k = kTs$, with $T_s$ being the sampling time and $k \in \mathbb{N}_{>0}$, as follows
\begin{subequations}\label{eq:NMPC_formulation}
    \begin{align}
    &\minimize_{\bar{\mathbf{x}}, \, \bar{\mathbf{u}}, \, \mathbf{s} } \;\;
    { \sum\limits_{k=0}^N \lVert \mathbf{y}_{\mathrm{d},k} - \mathbf{y}_k \rVert^2_{\mathbf{Q}_1} + \lVert \mathbf{s}_{j,k} \rVert^2_{\mathbf{Q}_2} } \label{subeq:objectiveFunction} \\
    %
    &\quad \text{s.t.}~\; \bar{\mathbf{x}}_0 = \bar{\mathbf{x}}(\mathbf{t}_k), k = 0,  \label{subeq:stateEquation} \\
    &\;\;\; \quad \quad \bar{\mathbf{x}}_{k+1} = \mathbf{f}(\bar{\mathbf{x}}_k, \bar{\mathbf{u}}_k), k \in \{0, N-1\} , \label{subeq:sysDynamic} \\
    &\;\;\; \quad \quad \mathbf{y}_k = \mathbf{h}(\bar{\mathbf{x}}_k, \bar{\mathbf{u}}_k, \mathbf{p}_{M_k}), k \in \{0, N\}, \label{subeq:outputMap} \\
    &\;\;\; \quad \quad\underline{\gamma} \leq \mathbf{u}_k \leq \bar{\gamma}, k \in \{0, N\}, \label{subeq:uBound} \\
    &\;\;\; \quad \quad \underline{\dot{\gamma}} \leq \bar{\mathbf{u}}_k \leq \bar{\dot{\gamma}}, k \in \{0, N-1\} \label{subeq:dotuBound}, \\
    &\;\;\; \quad \quad \lvert \prescript{C}{}{x_{M_k}} / \prescript{C}{}{z_{M_k}} \rvert \leq \tan{\alpha}_h, k \in \{0, N\} \label{subeq:imageHorizontal}, \\
    &\;\;\; \quad \quad \lvert \prescript{C}{}{y_{M_k}} / \prescript{C}{}{y_{M_k}} \rvert \leq \tan{\alpha}_v, k \in \{0, N\} \label{subeq:imageVertical}, \\
    &\;\;\; \quad \quad \lVert \mathbf{p}_k - \mathbf{p}_{{\mathrm{o}_j},k} \rVert^2 + \mathbf{s}_{j,k}^2 \geq \Gamma_j^2, \mathbf{s}_{j,k} > 0, \label{subeq:obstacleAvoidance} \\
    &\;\;\; \quad \quad k \in \{0, N\}, j \in \{1, O\}, \nonumber 
    \end{align}
\end{subequations}
where~\eqref{subeq:objectiveFunction} is the objective function,~\eqref{subeq:stateEquation} sets the initial state conditions,~\eqref{subeq:sysDynamic} and~\eqref{subeq:outputMap} express the discretized dynamic model for the~\ac{GTMR} and the output signals of the system, respectively, while actuator limits are embedded in~\eqref{subeq:uBound} and \eqref{subeq:dotuBound}. The hard constraints~\eqref{subeq:imageHorizontal} and~\eqref{subeq:imageVertical} ensure that the target remains in the camera~\ac{FoV} during the whole trajectory tracking problem. The soft constraint~\eqref{subeq:obstacleAvoidance} prevents the drone from colliding with the $j$-th obstacle by enforcing the vehicle to maintain a safety distance $\Gamma_j$. The slack variables $\mathbf{s} = [s_1 \cdots s_O]^\top \in \mathbb{R}^O$, with $\mathbf{s}_j$ denoting the $j$-th element of the vector $\mathbf{s}$, add a maneuverability room margin to the optimization problem by ensuring continuity of the solution when constraints may arise unfeasibility issues. Finally, the vectors $\bar{\mathbf{u}}_k$, $\bar{\mathbf{x}}_k$, $\mathbf{s}_{j,k}$, $\mathbf{y}_{\mathrm{d},k}$ and $\mathbf{y}_k$ denote the $k$-th element of vectors $\bar{\mathbf{u}}$, $\bar{\mathbf{x}}$, $\mathbf{s}_j$, $\mathbf{y}_{\mathrm{d}}$ and $\mathbf{y}$, respectively.

Similarly to~\cite{Castillo2018MED, BarrosIFAC2020}, this approach allows relaxing the constraint on the obstacle avoidance to guarantee feasible solutions in tight situations by introducing a penalty term in the objective function~\eqref{subeq:objectiveFunction}, i.e., $\lVert \mathbf{s}_j \rVert^2_{\mathbf{Q}_2}$. Specifically, when the~\ac{GTMR} gets too close to the obstacles, violating the safety distance constraint~\eqref{subeq:obstacleAvoidance}, then the slack variable $\mathbf{s}_j$ takes a positive value to satisfy the inequality~\eqref{subeq:obstacleAvoidance}, but this also adds a penalization term in the cost function~\eqref{subeq:objectiveFunction}.

\displayRemarks[SoftConstraints]{Soft Constraints}: Even though soft constraints are designed to be possibly violated, such as in~\eqref{subeq:obstacleAvoidance}, this situation should be minimized. This can be done by properly tuning the weighted matrix $\mathbf{Q}_2$. This approach allows flexible obstacle definition, guaranteeing feasible solutions.




\section{Simulation Results}
\label{sec:simulationResults}

In this section we provide MATLAB simulation results to validate the proposed control strategy. Specifically, the optimal control problem was coded using the MATMPC framework~\cite{Chen2019ECC}, with a $4$\textsuperscript{th} fixed step Runge-Kutta integrator and sampling time $T_s = \SI{15}{\milli\second}$, and qpOASES~\cite{Ferreau2014} as solver. All simulations were performed on a laptop with a i7-8565U processor ($\SI{1.80}{\giga\hertz}$) and $32$GB of RAM running on Ubuntu 20.04. Animation videos with the results of the numerical simulations are available at~\url{http://mrs.felk.cvut.cz/perception-aware-nmpc}.

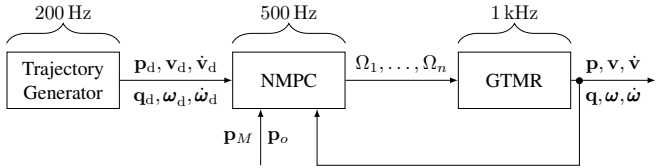
\begin{figure}[tb]
    \centering
	\scalebox{0.75}{
		\begin{tikzpicture}
		
		\node (TrajectoryGenerator) at (-4,0) [draw, rectangle, text centered, minimum width=1.5cm, 
		minimum height=1cm, text width=5em]{Trajectory\\Generator};
		
		\node (NMPC) at (0,0) [draw, rectangle, text centered, minimum 
		width=1cm, minimum height=1cm, text width=5em]{NMPC};
		
		\node (GTMR) at (4,0) [draw, rectangle, text centered, minimum 
		width=1cm, minimum height=1cm, text width=5em]{GTMR};
		
		\draw[-latex] (TrajectoryGenerator) -- node[above]{$\mathbf{p}_\mathrm{d}, \mathbf{v}_\mathrm{d}, \dot{\mathbf{v}}_\mathrm{d}$} node[below]{${\mathbf{q}_\mathrm{d}, \pmb{\omega}}_\mathrm{d}, \dot{\pmb{\omega}}_\mathrm{d}$} (NMPC);
		\draw[-latex] (NMPC) -- node[above]{$\Omega_1, \dots, \Omega_n$} (GTMR);
		\draw[-latex] (GTMR) -- node[above]{$\mathbf{p}, \mathbf{v}, \dot{\mathbf{v}}$} node[below]{${\mathbf{q}, \pmb{\omega}}, \dot{\pmb{\omega}}$} ($(GTMR) + (2.5, 0) $);
		\draw[-latex] ($(NMPC.south) - (0.5, 1.0)$) -- node[right]{$\mathbf{p}_o$} node[left]{$\mathbf{p}_M$} ( $(NMPC.south) - (0.5, 0)$ );
		\draw[-latex] ($(GTMR.south) + (1.15,0.5) $) -- ( $(GTMR.south) - (-1.15,1)$ ) -- ( $(GTMR.south) - (3.5,1)$ ) -- ( $(GTMR.south) - (3.5,0)$ ); 
		\node at  ($(GTMR.south) + (1.15,0.5) $) [circle,fill,draw,inner sep=0pt,minimum size=3pt, text centered]{};
		
		\draw [decorate, decoration={brace, amplitude=10pt, mirror}, xshift=0pt, yshift=0pt] ($ (TrajectoryGenerator.north east) + (0,0.15)$ ) -- ($ (TrajectoryGenerator.north west) + (0,0.15)$) node [black,midway,xshift=0cm, yshift=0.55cm] {$\SI{200}{\hertz}$};
		\draw [decorate, decoration={brace, amplitude=10pt, mirror}, xshift=0pt, yshift=0pt] ($ (NMPC.north east) + (0,0.15)$ ) -- ($ (NMPC.north west) + (0,0.15)$) node [black,midway,xshift=0cm, yshift=0.55cm] {$\SI{500}{\hertz}$};
		\draw [decorate, decoration={brace, amplitude=10pt, mirror}, xshift=0pt, yshift=0pt] ($ (GTMR.north east) + (0,0.15)$ ) -- ($ (GTMR.north west) + (0,0.15)$) node [black,midway,xshift=0cm, yshift=0.55cm] {$\SI{1}{\kilo\hertz}$};
		
		\end{tikzpicture}
	}
    \caption{Block diagram of the proposed optimal control strategy.}
    \label{fig:controlArchitecture}
\end{figure}

The control architecture is shown in Fig.~\ref{fig:controlArchitecture}. A reference generator working at $\SI{200}{\hertz}$ provides the reference trajectory ($\mathbf{p}_\mathrm{d}^\top, \mathbf{q}_\mathrm{d}^\top, \mathbf{v}_\mathrm{d}^\top, \bm{\omega}_\mathrm{d}^\top, \dot{\mathbf{v}}_\mathrm{d}^\top, \dot{\bm{\omega}}_\mathrm{d}^\top$) to the~\ac{NMPC} which runs at $\SI{500}{\hertz}$. The output of the~\ac{NMPC} are the propeller spinning velocities $\bm{\Omega}$ supplied to the aerial vehicle to control its motion. The high frequency at which the control algorithm works simulates the times required in real applications to control the vehicle dynamics. The optimal control strategy (Section~\ref{sec:optimalControlProblem}) runs considering $N=50$ shooting points and a prediction horizon of $\SI{0.75}{\second}$, while the system dynamics~\eqref{eq:modelEquations} are integrated with a sampling time of $\SI{1}{\milli\second}$ Table~\ref{tab:controlParameters} reports the~\ac{NMPC} gains along with the~\ac{GTMR}'s parameters values. The parameters provide a balanced trade-off between efficiency, safety and tracking.

\begin{table}
	\centering
	\begin{adjustbox}{max width=1\columnwidth}
		\begin{tabular}{|c|c|c|c|c|c|c|c|}
		    \hline
		    \textbf{Sym.} & \textbf{Value} & \textbf{Sym.} & \textbf{Value} & \textbf{Sym.} & \textbf{Value} & \textbf{Sym.} & \textbf{Value} \\
			\hline
			\hline
			$Q_{p}$ & $\num{1}$ & $Q_{q}$ & $\num{1}$ & $Q_{v}$ & $\num{0.1}$ & $Q_{\omega}$ & $\num{0.1}$ \\
			$Q_{\dot{v}}$ & $\num{0.01}$ & $Q_{\dot{\omega}}$ & $\num{0.01}$ & $Q_{c{\beta}}$ & $\num{100}$ & $Q_{\dot{c{\beta}}}$ & $\num{100}$ \\
			$Q_{d_\mathrm{trg}}$ & $\num{10}$ & $Q_{s}$ & $\num{1e4}$ & $\bar{\gamma}$ & $\SI{40}{\hertz}$ & $\underline{\gamma}$ & $\SI{90}{\hertz}$ \\
			$\bar{\dot{\gamma}}$ & $\SI{200}{\hertz\per\second}$ & $\underline{\dot{\gamma}}$ & $\SI{-110}{\hertz\per\second}$ &  $m$ & $\SI{1.042}{\kilogram}$ & $g$ & $\SI{9.84}{\meter\per\square\second}$ \\
			$\mathbf{J}_1$ & $\num{0.015}$ & $\mathbf{J}_2$ & $\num{0.015}$ & $\mathbf{J}_3$ & $\num{0.070}$ & $\Gamma$ & $\SI{1}{\meter}$ \\
			$\alpha_h$ & $\pi/2$ & $\alpha_v$ & $\pi/2$ & $\Upsilon$ & $\SI{1}{\meter}$ & - & -\\
			\hline
		\end{tabular}
	\end{adjustbox}
	\caption{List of parameters and their values. $Q_k$ refers to the $k$-th element of the weighted diagonal matrix $\mathbf{Q}_1$ and $\mathbf{Q}_2$~\ac{wrt} the output map $\mathbf{y}$ and the slack variables $\mathbf{s}$, respectively.} 
	\label{tab:controlParameters}
\end{table}

The trajectory tracking scenario consists of a target following an ascending ramp trajectory at a constant velocity of $\lvert \mathbf{v}_\mathrm{d} \rvert = \SI{1}{\meter\per\second}$ for $\SI{10}{\second}$. A single integrator model was used to characterize the motion of the target. This model is a canonical example of a first order control system. As described in~\cite{Zhao2017ICCA}, demonstrating numerical convergence with this model guarantees convergence even with more complex models, as long as the control law is suitably adapted to the various motion constraints. 

\begin{figure}[tb]
    \centering
    \vspace{0.5em}
    \hspace{-1.25em}
    \adjincludegraphics[width=0.5\textwidth, trim={{0.1\width} {0.15\height} {0.0\width} 
		{0.2\height}}, clip]{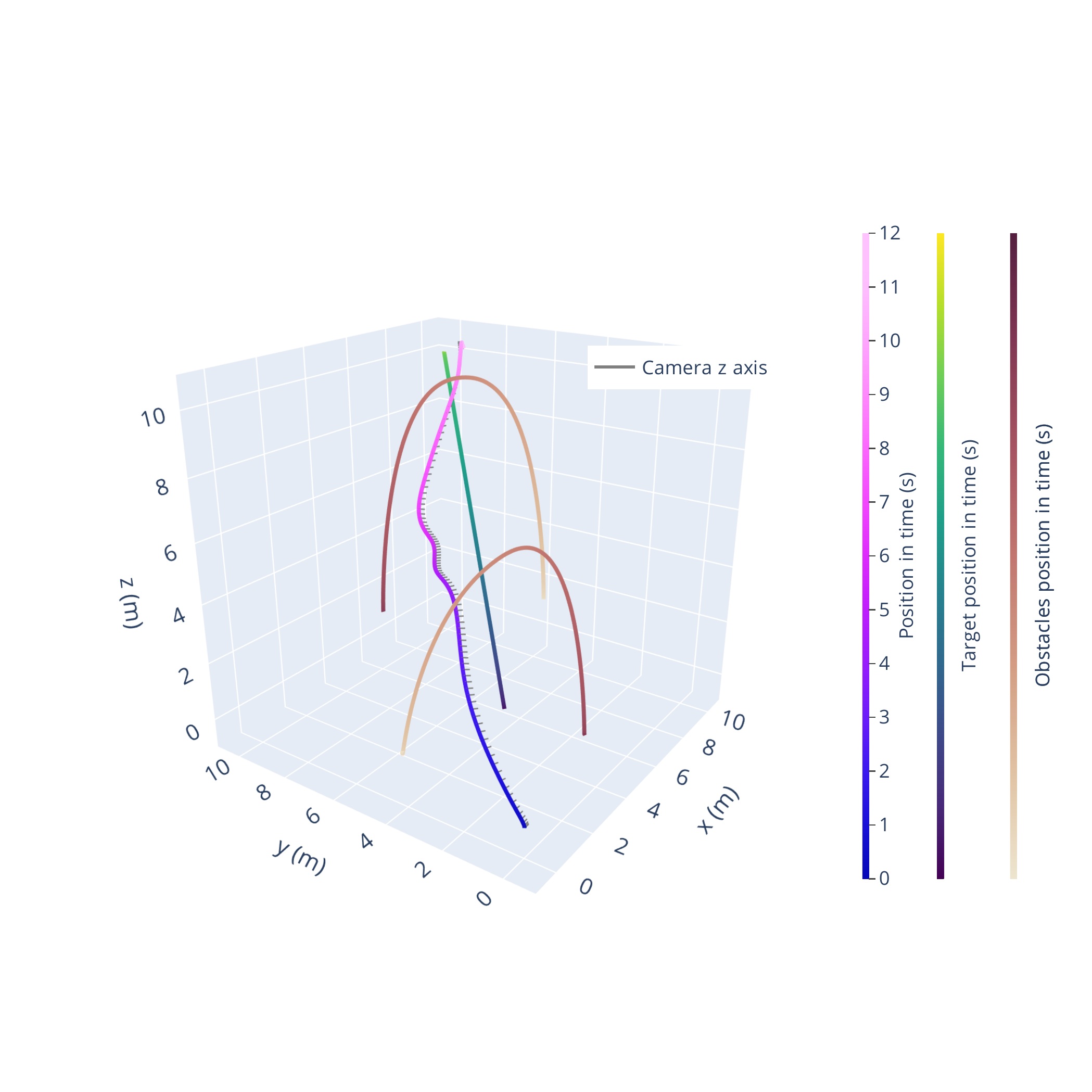}
    \caption{The vehicle motion in the trajectory tracking and collision avoidance scenario. Gradient colors show the drone, target and obstacles motion during time. A short segment represents the $z$-axis orientation of the camera ($\hat{z}_c$).}
    \label{fig:simulationResults3D}
\end{figure}

The~\ac{GTMR} is equipped with $n=4$ propellers arranged parallel to the $xy$-plane of the body frame $\mathcal{F}_B$ and having the same direction of its $z$-axis (i.e., a coplanar under-actuated platform). The initial drone and target positions in the world frame $\mathcal{F}_W$ are $\mathbf{p}=[0, 0, 0]^\top$ and $\mathbf{p}_M = [6, 6, 0]^\top$, respectively. Two dynamic obstacles have been considered populating the workspace area with initial position $\mathbf{p}_{\mathrm{o}_1} = [2, 6, 0]^\top$ and $\mathbf{p}_{\mathrm{o}_2} = [10, 6, 2]^\top$. The origin of the visual generic sensor $O_C$ lies on the vehicle body frame $\mathcal{F}_B$ and is positive translated $\SI{0.1}{\meter}$ along its $x$-axis ($\hat{x}_B)$. The overall scenario is depicted in Fig.~\ref{fig:simulationResults3D} along with the vehicle, obstacles, and target motion during time.

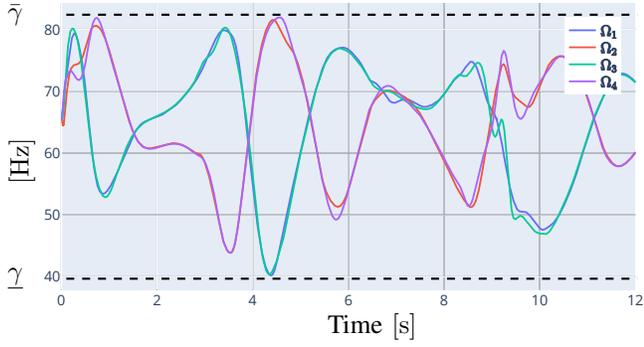
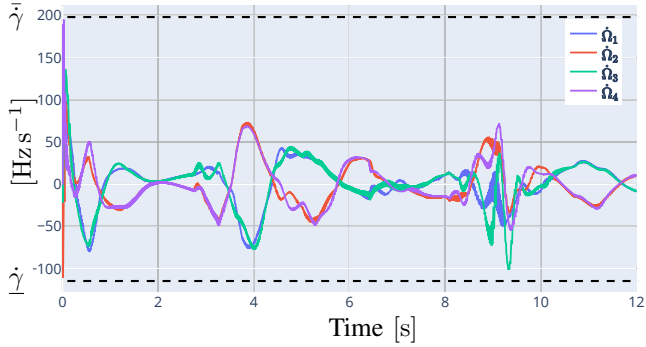
\begin{figure}[tb]
    \hspace{-0.1cm}
    \begin{subfigure}{0.95\columnwidth}
        \vspace{1em}
        \begin{tikzpicture}
	        \node[anchor=south west,inner sep=0] (img) at (0,0) { 
	        \adjincludegraphics[width=1.10\textwidth, trim={{0.06\width} {0.1\height} {0.0\width} {0.15\height}}, clip]{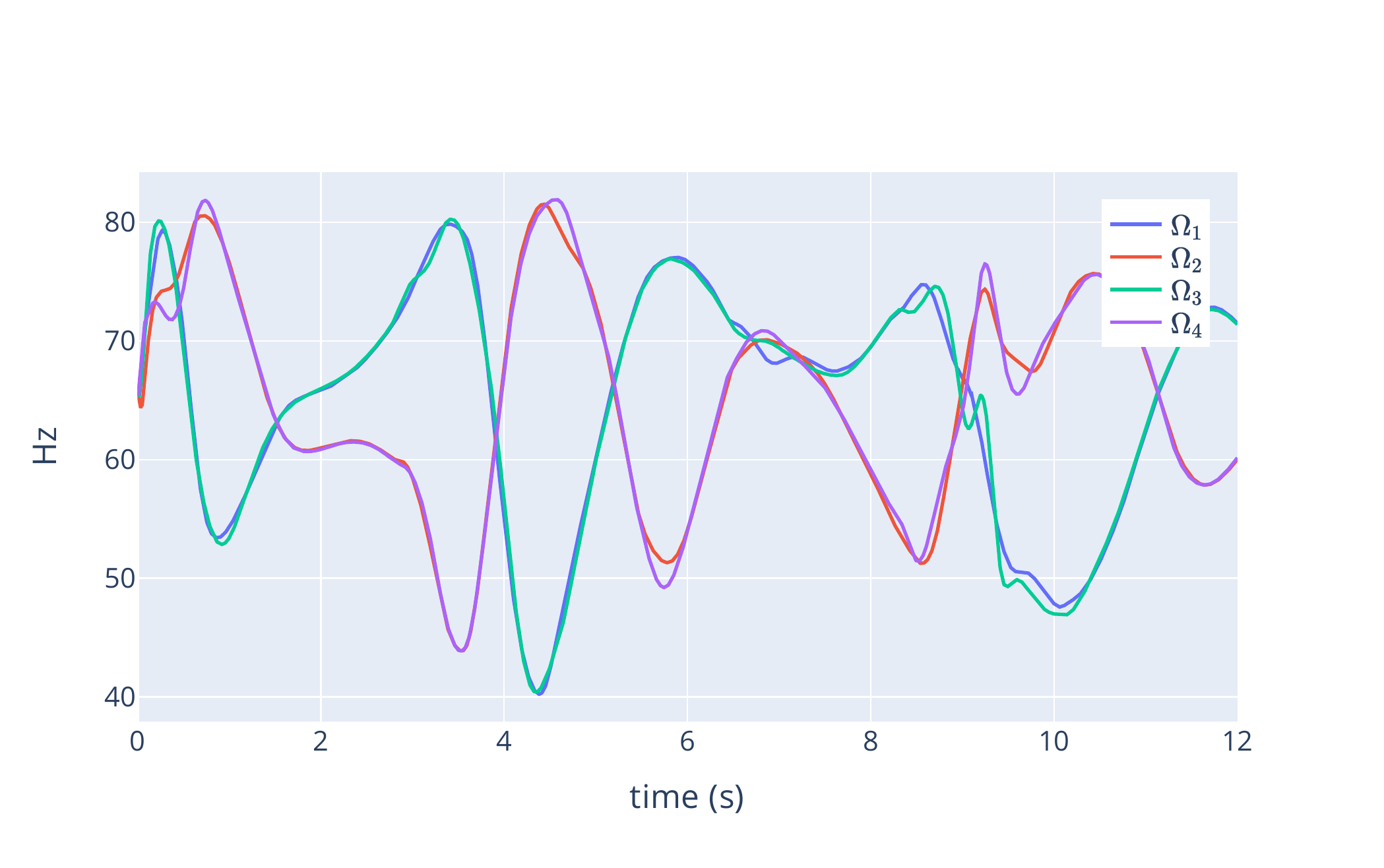}};
	        \begin{scope}[x={(img.south east)},y={(img.north west)}]
	          \draw[black, dashed, thick] (0.05, 0.9) -- (0.89,0.9);
	          \draw[black, dashed, thick] (0.05, 0.125) -- (0.89,0.125);
	          \draw (0, 0.9) node[left]{$\bar{\gamma}$};
	          \draw (0, 0.125) node[left]{$\underline{\gamma}$};
	          \draw (-0.015,0.375) node[right, rotate=90]{$[\si{\hertz}]$};
	          \draw (0.5,0.05) node[below]{Time $[\si{\second}]$};
	        \end{scope}
	    \end{tikzpicture}
	    \vspace{-0.5cm}
		\caption{Propeller angular velocities $\Omega_i$, with $i=\{1, 2, 3, 4\}$.}
		\label{fig:angularVelocities}
    \end{subfigure}
    \\
    \vspace{-0.05cm}
    \begin{subfigure}{0.95\columnwidth}
        \hspace{-0.1cm}
        \begin{tikzpicture}
	        \node[anchor=south west,inner sep=0] (img) at (0,0) { 
	        \adjincludegraphics[width=1.1\textwidth, trim={{0.06\width} {0.1\height} {0.0\width} {0.15\height}}, clip]{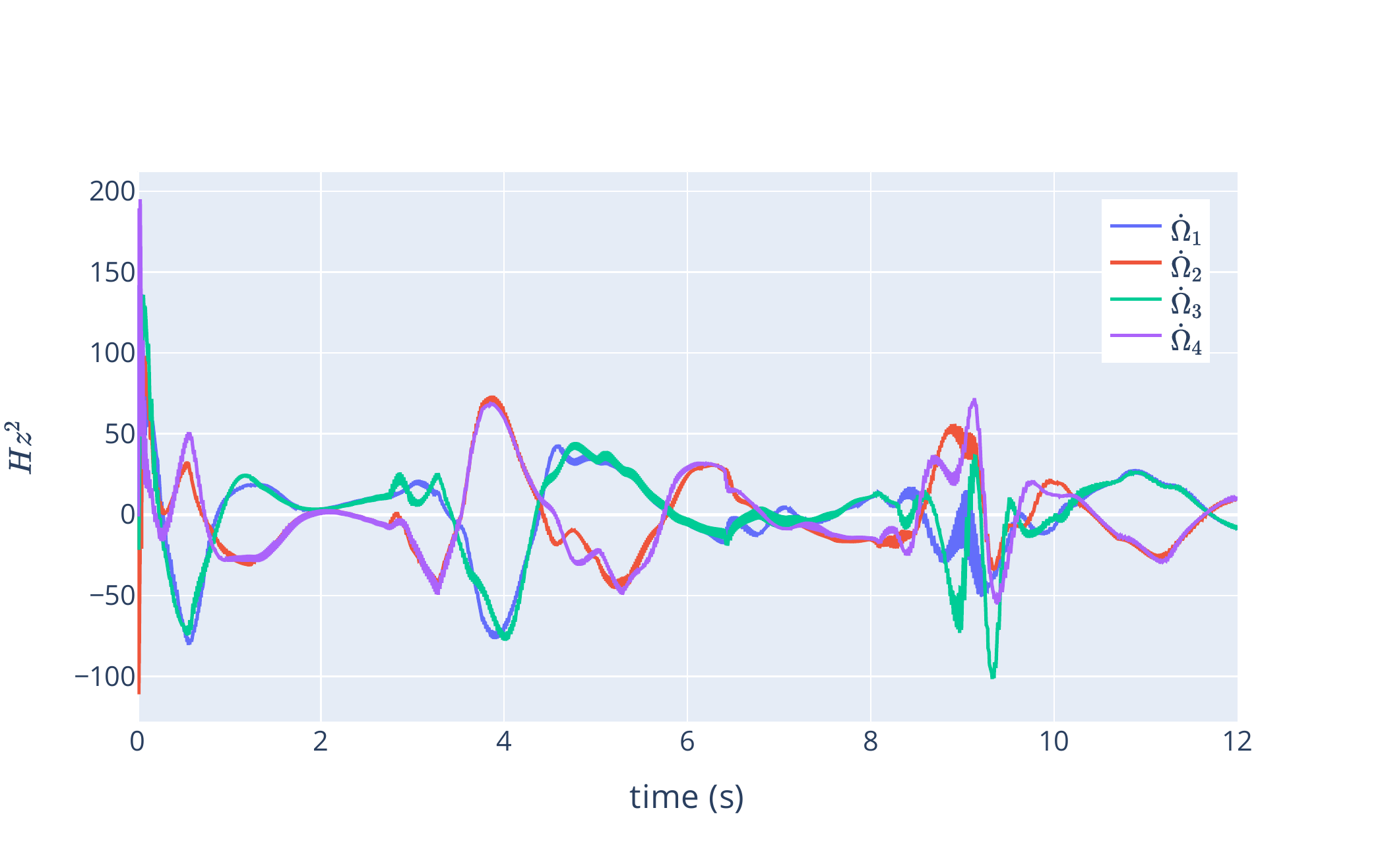}};
	        \begin{scope}[x={(img.south east)},y={(img.north west)}]
	          \draw[black, dashed, thick] (0.05, 0.9) -- (0.89,0.9);
	          \draw[black, dashed, thick] (0.05, 0.125) -- (0.89,0.125);
	          \draw (0, 0.9) node[left]{$\bar{\dot{\gamma}}$};
	          \draw (0, 0.125) node[left]{$\underline{\dot{\gamma}}$};
	          \draw (-0.015,0.375) node[right, rotate=90]{$[\si{\hertz\per\second}]$};
	          \draw (0.5,0.05) node[below]{Time $[\si{\second}]$};
	        \end{scope}
	    \end{tikzpicture}
	    \vspace{-0.5cm}
		\caption{Control input variables $\bar{\mathbf{u}}=[\dot{\Omega}_1 \, \dot{\Omega}_2 \, \dot{\Omega}_3 \, \dot{\Omega}_4]^\top$.}
		\label{fig:variationAngularVelocityOverTime}
    \end{subfigure}
    \caption{Propeller spinning velocities and their variation during time along with the corresponding bounds $\bar{\bm{\gamma}}$, $\underline{\bm{\gamma}}$, $\dot{\bar{\bm{\gamma}}}$, and $\dot{\underline{\bm{\gamma}}}$. }
    \label{fig:propellerAngularVelocitiesAndVariation}
\end{figure}

\begin{figure}[tb]
    \vspace{2em}
    \centering
    \begin{tikzpicture}
        \hspace{-1.25em}
        \node[anchor=south west,inner sep=0] (img) at (0,0) { 
        \adjincludegraphics[width=0.475\textwidth, height=0.2\textheight, trim={{0.06\width} {0.15\height} {0.075\width} {0.245\height}},clip]{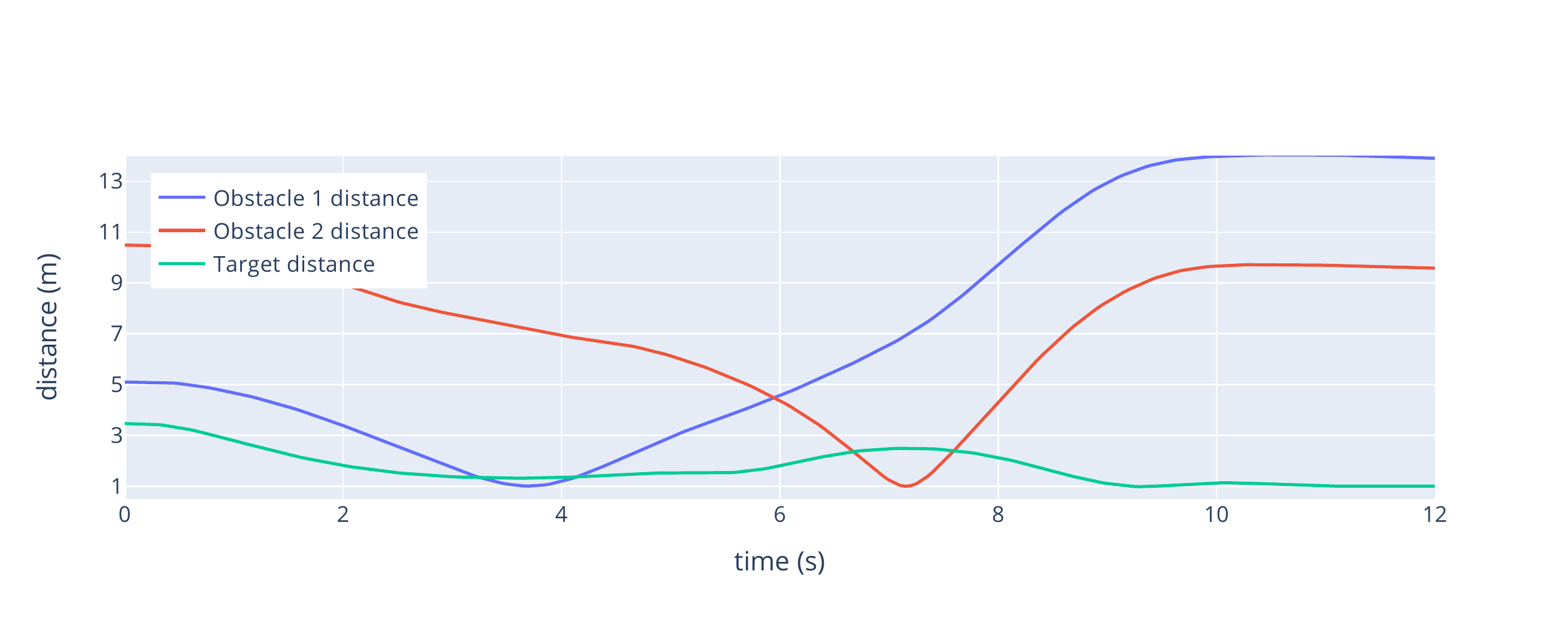}};
        \begin{scope}[x={(img.south east)},y={(img.north west)}]
          \draw (-0.02,0.45) node[right, rotate=90]{$[\si{\meter}]$};
          \draw (0.5,0.05) node[below]{Time $[\si{\second}]$};
        \end{scope}
    \end{tikzpicture}
    \vspace{-0.7cm}
    \caption{Distances between the~\ac{GTMR} and the obstacles (in blue and red) and the~\ac{GTMR} and the target (in green) during time.}
    \label{fig:obstacleDistance}
\end{figure}

Figs~\ref{fig:angularVelocities} and~\ref{fig:variationAngularVelocityOverTime} show the evolution of the propeller spinning velocities $\bm{\Omega}$ and their variation $\dot{\bm{\Omega}}$ during time. As it can be seen from the figures, both values remain within the boundaries ($\underline{\gamma}$, $\bar{\gamma}$, $\dot{\underline{\gamma}}$, $\dot{\bar{\gamma}}$) as required by the constraints~\eqref{subeq:uBound} and~\eqref{subeq:dotuBound}. Figure~\ref{fig:obstacleDistance} shows the distances between the target ($\mathbf{p}_M$) and the obstacles ($\mathbf{p}_{\mathrm{o}_1}$ and $\mathbf{p}_{\mathrm{o}_2}$)~\ac{wrt} the drone's position ($\mathbf{p}$). As it can be seen from the graph, the collision avoidance avoidance constraints~\eqref{subeq:obstacleAvoidance} is always satisfied for both for the obstacles and the target~\eqref{subeq:objectiveFunction}.

\displayRemarks[PenaltySoftConstraint]{Penalties on Soft Constraints}: It is worth noticing that, the effort required to tune the penalty values $\mathbf{Q}_2$ on the slack variables $\mathbf{s}$ directly affects the amount of constraints violation. Roughly speaking, imposing very strict penalties increases the difficulty in driving the solution towards the optimum while distancing the vehicle from the lower bound. Conversely, if the penalty is not strict enough, then the search will tend to stall outside the feasible region and thereby violate the constraint. Hence, such an approach does not guarantee that the violation will not occur. 

\section{Conclusions}
\label{sec:conclusions}

In this paper, a perception-aware~\ac{NMPC} strategy for vision-based target tracking and obstacle avoidance have been proposed. In particular, an optimization problem was set to control the behavior of a multi-rotor aerial vehicle enforcing the visibility coverage of a target while accounting for real actuation limits and obstacles avoidance. A full nonlinear generic constrained model has been considered for the flight control system design covering both coplanar under-actuated platforms and fully-actuated tilted propellers. Numerical simulations achieved in MATLAB have demonstrated the feasibility of the proposed control approach, aiming towards the fulfillment of real-word tests. Future work includes relaxing the assumptions on the obstacle motion over the prediction horizon of the optimization problem. In addition, more challenging scenarios will be investigated, such the combination of static and dynamic obstacles, in the direction of field experiments. 



\bibliographystyle{IEEEtran}
\bibliography{bib_short}

\end{document}